\documentclass{llncs}

\usepackage{makeidx}  

\usepackage{graphicx}

\begin{document}

\title{\Large Variability of Behaviour in Electricity Load Profile Clustering; Who Does Things at the Same Time Each Day?}

\titlerunning{Electricity Behaviour Variability}

\author{Ian Dent\inst{1} \and Tony Craig\inst{2} \and
Uwe Aickelin\inst{1} \and Tom Rodden\inst{1}}
\authorrunning{Ian Dent et al.} 
\institute{School of Computer Science, University of Nottingham, Nottingham NG8 1BB, UK,\\
\email{psxid@nottingham.ac.uk},\\ WWW home page:
\texttt{http://ima.ac.uk/dent}
\and
The James Hutton Institute, Aberdeen, UK}

\maketitle
 

\begin{abstract} 

UK electricity market changes provide opportunities to alter households' electricity usage patterns for the benefit of the overall electricity network. Work on clustering similar households has concentrated on daily load profiles and the variability in regular household behaviours has not been considered. Those households with most variability in regular activities may be the most receptive to incentives to change timing.

Whether using the variability of regular behaviour allows the creation of more consistent groupings of households is investigated and compared with daily load profile clustering.  204 UK households are analysed to find repeating patterns (motifs). Variability in the time of the motif is used as the basis for clustering households. Different clustering algorithms are assessed by the consistency of the results.

Findings show that variability of behaviour, using motifs, provides more consistent groupings of households across different clustering algorithms and allows for more efficient targeting of behaviour change interventions.

\end{abstract}

\section{Background and Motivation}

The electricity market in the UK is undergoing dramatic changes. Legal, social and political drivers for a more carbon efficient electricity network, along with the dramatically increased flow of data from households through the deployment of smart meters, requires a transformation of existing practices. In particular, the change of the frequency of sampling of electricity usage, by using smart meters, alters the level of understanding of households' behaviour that is possible \cite{Energy2009}.

One approach to address the pressures on the electricity network is the application of Demand Side Management (DSM) techniques to achieve changes in consumer behaviour. DSM is defined as ``systematic utility and government activities designed to change the amount and/or timing of the customer's use of electricity'' for the collective benefit of society, the utility company, and its customers \cite{river2005primer}. The peak time for electricity usage in the UK is during the early evening and the successful application of techniques to reduce, or move, the peak usage would improve the overall efficiency of the electricity network. 

To allow selection of appropriate DSM interventions, a good understanding of the existing behaviour of households is needed. Firstly, knowledge is needed on an individual household that can be deduced from house-wide electricity metering. Secondly, a method is required to group large numbers of households into a manageable number of archetypal groups where the members display similar characteristics. This approach allows for cost effective targeting of the most appropriate subset of customers whilst allowing the company management to deal with a manageable number of archetypes \cite{mooi2011concise}. 

There is an extensive body of work on clustering households which includes comparing or combining timed meter readings to create additional attributes that contribute to the quality of the clustering \cite{ramos2007knowledge}. However, little work has focused on how the daily activity patterns of the household vary from day to day and how this can be used for clustering. For instance, some households will be creatures of habit and will eat their evening meal at almost the same time each evening, whilst others have a much more variable activity pattern and will eat at different times. Elleg\.{a}rd and Palm  \cite{ellegaard2011visualizing} have investigated the variability of behaviour using diaries and interviews but have not used analysis of meter data.

Clustering households using their degree of variability in behaviour, as shown by electricity consumption, provides a way of identifying the subset of electricity users who may be most receptive to an intervention to influence their activity patterns. The intervention may be to reward households for NOT changing their current pattern of usage if it is already as desired by the utility company.

This paper addresses the question of whether making use of the variability of behaviour (as shown by the electricity meter data) provides ``better'' groupings of households for the purpose of DSM than those provided by using daily load profiles. The judgement of ``better'' is measured by implementing a number of different clustering techniques and measuring the degree of overlap between the clusters found. A consistent set of clusters across the different clustering algorithms implies a better, and more useful, approach to generating the clusters.

The investigation of household electricity load profiles is an important area of research given the centrality of such patterns in directly addressing the needs of the electricity industry, both now and in the future. This work extends existing load profile work by taking electricity meter data streams and developing new ways of representing the household that can be used as the basis for clustering using existing data mining techniques. The identification of repeating motifs and the investigation of how the timing of the motifs varies from day to day, as a key behavioural trait of the household, is a novel area of research. An improvement in creating useful archetypes can have major financial and environmental benefits.
 
\section{Methods and Technical Solutions}

\subsection{Load Profiling}
\label{loadprofile}

There has been extensive research on determining daily load profiles to represent a household's electricity usage \cite{chicco2012overview}. In many cases, (e.g., \cite{ramos2012typical}), the daily load profiles are used as the basis for clustering ``similar'' households together to develop a small set of archetypal profiles which can be used for targeting of behaviour change interventions. Previous work has used different clustering techniques with the majority of the published literature using hierarchical clustering. 

The common approach is to define a subset of the data (e.g. by season and/or by day of the week) and then to create average daily profiles for a household from the electricity meter data. The shapes of these daily profiles are then clustered to group similar shapes together. A representative profile is defined (e.g. by averaging all the members of the cluster) to produce a archetypal daily load profile for that cluster of households.

Previous work has not investigated how households may exhibit different behaviour from day to day and how these differences may be used as a distinguishing feature of the household and a basis for clustering.

\subsection{Motifs}
\label{motif}

The electricity meter data reading stream from a household can be plotted as a graph of usage against time and regular activities appear as similar shaped patterns. Short patterns that repeat are defined as ``motifs'' and detection of these motifs, and their timing, can inform understanding of household behaviour.

This work uses the SAX (Symbolic Aggregate approXimation) technique which allows symbolic representation of time series data \cite{lin2007experiencing,shieh2008sax}. Other motif finding algorithms could also be incorporated into the proposed approach to identify the flexibility of behaviour (e.g. \cite{mueen2009exact}). To assess variability within a household, it is necessary to detect the repeating motifs that are assumed to signify particular activities (e.g., cooking the evening meal). These are generally of a similar shape on different days but show some differences due to noise caused by other activities within the household (e.g., a fridge automatically running). The SAX approach of symbolising the real valued meter readings is useful as it allows for approximate matching (as various ranges of readings map to a single symbol).

Lines et al \cite{Lines2011} applies motif finding to UK data to detect the use of particular appliances, drawn from a set of known appliances. This contrasts with the focus in this paper which is to find interesting, repeating patterns of behaviour without the need to define the activity that the motif represents. Appliances that can be consistently and accurately detected can be used with the approach detailed here by extending the analysis of the timing of repeating motifs to the analysis of variability of timing of appliance usage.

\subsection{North East Scotland Electricity Monitoring Project (NESEMP)}
\label{NESEMP}

This study makes use of data collected as part of the ongoing NESEMP which is examining the relationship between different types of energy feedback and psycho-social measures including individual environmental attitudes, household characteristics, and everyday behaviours.  As part of this ongoing project, several hundred households are being monitored and the electricity usage is recorded every five minutes using CurrentCost monitors \cite{craig2014north}. 

After removing data for households with insufficient readings, the data is loaded into a MySQL database and the readings are aligned with exact 5 minute boundaries (e.g. 1pm, 1.05pm, etc.) by interpolation between the actual readings. This is achieved by calculating the reading at an exact 5 minute point (e.g. 1.05pm) by considering the actual readings before and after that time and by calculating the reading such that the total usage over a longer period is the same whether the interpolated readings or the original actual readings are used \cite{dent2012}. This results in a set of 288 readings (one for every 5 minute period in the day) for each of the households in the database.

Each day of sampling is labelled in a number of ways such as ``working day'' or ``summer''  to aid selection of particular subsets of data.

\subsection{Detecting Motifs}
\label{motifs}

\begin{figure}[h]
\centering
\includegraphics[width=0.95\columnwidth]{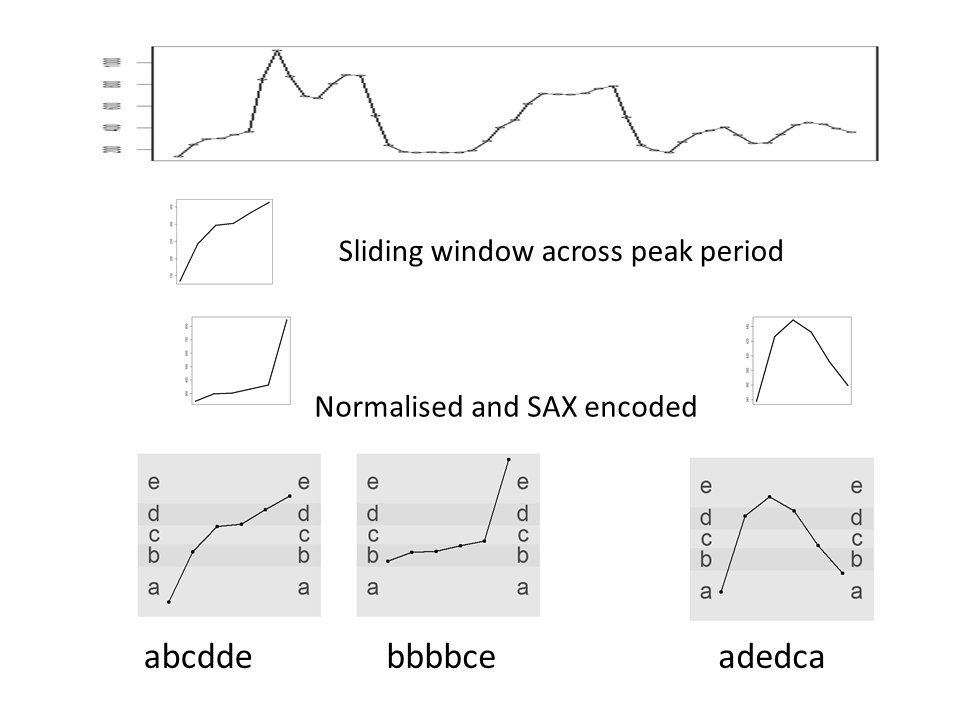}
\caption{Example of symbolisation (alphabet of 5, motif length of 6)}
\label{sax-explanation}
\end{figure}

To find motifs within the data, each period of interest within the day (e.g., the peak period) for each household is examined by taking a moving window over the period. The subset of the meter readings within the moving window is then converted into a string and stored. Next, the window moves on by one time period (5 minutes) and the conversion into a string is repeated. Using an alphabet size of 5 and a motif size of 6 (i.e., 30 minutes), analysing the 4pm to 8pm period provides a total of 49 x 5 minute readings for each day. As the interest is in changes in usage rather than absolute usage, these readings are compared with adjacent readings in time to produce 48 values (one for each 5 minute period) representing the change in usage since the last 5 minute reading. This results in 42 motifs stored for each day for each household (one for each possible 30 minute period within the peak time). Fig. \ref{sax-explanation} shows an example of how the symbolised motifs are built up. The top graph shows the 5 minute readings for the 4 hour peak period. A sliding window of 6 readings (30 minutes) is taken across the peak period with the first 2 and the last window shown. Each window is normalised within the values in the window and then translated into the symbolised representations as shown at the bottom of the diagram.

The analysis uses an alphabet of 5 symbols (i.e., the letters ``a'' to ``e'') to represent the motifs. 5 is selected as a reasonable compromise between having too few symbols, and thus not detecting changes in electricity consumption, and having too many and thus generating too many patterns that do not repeat. The symbolisation translates readings within a particular range into a given letter and thus similar, although not identical, readings are translated into the same letter. The resulting motifs for 2 windows may be identical whereas the original readings may only be approximately similar.

The motif size selected is 6 corresponding to a 30 minute (i.e., 6 x 5 minutes) period. This figure was selected as the UK electricity settlement market uses a 30 minute period \cite{Elexon2012} and 30 minutes is also a reasonable period that will allow time for activities such as showering. 

The motifs are built from the graph shape without regard to absolute value of the data.  A possible effect of this is to find motifs within what is the general noise associated with the meter readings. This is avoided by ignoring any motifs within a window which have a range of less than 100W.

As the motifs are created by shifting a moving window over the stream of data, overlapping periods are considered and periods with no activity except for one change in meter reading will lead to a series of motifs that are similar. For example, a long period of no activity except for a jump of +200W will lead to motifs being found such as ccccca, ccccac, cccacc, etc. As only one of these is interesting for further analysis, the others are excluded.

The top motif (the one that occurs most often within a household) is further examined for the times when the motif occurs on each day. The number of times the motif occurs, and the standard deviation of the time of occurrence, are calculated for each household. Similarly, the second and third most common motifs within a household are identified and the variability in timing calculated.

Other useful measures relating to the motifs found within a household are also calculated including the number of different motifs (occurring at least twice) and the number of different motifs occurring on at least 30\% of the days sampled for the household. The 30\% figure is selected as a reasonable number to ensure only regularly repeating patterns are considered.

The attributes calculated for each household and used as input to the clustering algorithms are:
\begin{enumerate}
\item Number of occurrences of the motif occurring most frequently during the peak period.
\item Variability in timing of the occurrence of the most frequent motif within the household. This is represented by the standard deviation of the timing (measured in minutes) around the mean start time. 
\item Number of occurrences of the second most frequent motif.
\item Variability in timing of the second most frequent motif.
\item Number of occurrences of the third most frequent motif.
\item Variability in timing of the third most frequent motif.
\item Total number of motifs for the household that occur at least twice. 
\item Total number of different motifs that occur on at least 30\% of days.
\end{enumerate}

\subsection{Clustering Algorithms}

Various clustering techniques are selected for evaluation of the different approaches to analysing the data. Note that, whilst possibly a useful additional benefit, this work does not focus on selecting the ``best'' clustering algorithm but uses a selection of algorithms to assess the benefits or otherwise of making use of the motif variability information. 

Based on the review by Chicco \cite{chicco2012overview} the following clustering algorithms are selected as the most commonly used in previous work: 
\begin{enumerate}
\item Kmeans is a well known algorithm that occurs in a number of examples of previous load profiling work. The algorithm requires a number of clusters (k) and works by randomly selecting an initial k locations for the centres of the clusters. Each data point is then assigned to one of the clusters by selecting the centre nearest to that data point. Once all the data points are assigned, each collection of points is considered, the new centre of the allocated points is calculated and the centre for that cluster is reassigned. The points are then reallocated to their new nearest centre and the algorithm continues until no changes are made to the allocations of points for an iteration \cite{jain1988algorithms}.
\item Fuzzy c means. This provides an extension of the kmeans algorithm allowing partial membership to more than one cluster. The algorithm provides additional output showing the degree of membership that each household has of each of the derived clusters \cite{bezdek1981}. For this analysis, each household is assigned to the cluster for which they have the highest degree of membership.
\item Self Organising Maps. The Self Organising Map (SOM) is a neural network algorithm that can be used to map a high dimension set of data into a lower dimension representation. In this paper, the mapping is to a 2 dimensional set of representations which are arranged in a hexagonal map. Each sample (e.g., the average load profile for a given household) is assigned to a position in the map depending on the closeness of the sample to the existing nodes assigned to each position in the map (using a Euclidean measure of distance). Initially the nodes are assigned at random but, over time, the map produces an arrangement where similar samples are placed closely together and dissimilar samples are placed far apart \cite{kohonen2002self}.
\item Hierarchical clustering. Most of the published load profiling work has used hierarchical clustering and this approach has the benefit of providing easily understood rules for cluster membership. The algorithm uses a dissimilarity matrix for the households and, starting initially with each household in its own cluster, proceeds by joining clusters which are most similar. The hierarchy is cut at a point to provide the desired number of clusters \cite{everitt2001cluster}. The Euclidean distance is used when creating the dissimilarity matrix and the Ward agglomeration method \cite{ward1963hierarchical} is used for combining clusters. The Ward method minimises the sum of squares of possible clusters when selecting households to combine. Other agglomeration techniques tend to create a few small clusters containing extreme valued households plus one large cluster containing the remainder of the households. 
\item Random Forests \cite{breiman2001random} is used to create a dissimilarity matrix which is used with Partitioning Around Medoids (pam) to form clusters. This is implemented using the R package randomForest \cite{Liaw2002}. 
\end{enumerate}

A common issue is the appropriate setting for the number of clusters. To match common practice within the electricity industry, 8 clusters are selected. The UK electricity industry has worked with 8 load profiles since the 1990s \cite{Association1997}. Figueiredo et al \cite{figueiredo2005electric} report that the Portuguese electricity utility aim for a number of clusters between 6 and 9.

\subsection{Cluster Validity Measures}

To assess the benefits of a particular cluster solution an appropriate cluster validity index needs to be used. Many have been considered in the literature with the Mean Index Adequacy (MIA) and the Cluster Dispersion Indicator (CDI) \cite{chicco2003customer} used in most of the published load profiling work. Lower values for the CDI and MIA measure denote ``better'' solutions. 

The data to be clustered consists of $M$ records numbered as $m=1,..M$. Each record has $H$ attributes numbered as $h=1,..H$. The $h$th attribute for the $i$th record is designated as $m_i(h)$.

The data is clustered into $K$ clusters (numbered as $k=1,..,K$). Each cluster has $R_{k}$ members where $r_{(k)}$ is the $r$th record assigned to cluster $k$ and $C_{(k)}$ is the calculated centre of the cluster $k$.

The distance ($d$) between 2 records is defined as:
\begin{equation}
d(m_{i},m_{j}) = \sqrt{ \frac{1}{H} \sum\limits_{h=1}^H (m_i(h) - m_j(h))^2}
\end{equation}
where $m_i(h)$ and $m_j(h)$ are the $h$th attributes for two records, $m_i$ and $m_j$.

The ``within set distance'' $\hat{d}(S)$ of the members of a set, $S$ with $N$ members ($s_j$ where $j=1,..,N$) is defined as:
\begin{equation}
\hat{d}(S) = \sqrt{ \frac{1}{2N} \sum\limits_{n=1}^{N} \sum\limits_{p=1}^{N} d^{2}(s_n, s_p)}
\end{equation}

The MIA gives a value which relies on the amount by which each cluster is compact - i.e., if the members in the cluster are close together the MIA is low.

\begin{equation}
MIA = \sqrt{ \frac{1}{K} \sum\limits_{k=1}^K \sum\limits_r d^2 (r_{(k)}, C_{(k)})}
\label{equation-eqn2}
\end{equation}

The CDI depends on the distance between the members of the same cluster (as for the MIA) but also incorporates information on the distances between the representative load diagrams (i.e., the centroids) for each cluster. This therefore measures both the compactness of the clusters and the amount by which each cluster differs from the others.

\begin{equation}
CDI = \frac{1}{\hat{d}(C)} \sqrt{ \frac{1}{K} \sum\limits_{k=1}^K \hat{d}^2(R_k)}
\label{equation-eqn3}
\end{equation}
where $C$ is the set of cluster centres and $R_{k}$ is the $k$th cluster members set.

\subsection{Processing}
\label{processing}

UK specific data is used to generate average daily load profiles for each household which are clustered to provide a baseline for comparison. Selected clustering algorithms are applied to the data and validity indexes are used to produce a measure of the quality of the partitions found. 

Next, the novel approach of identifying motifs within the data, and measuring the variability in timing of the motifs, is used to generate a new set of derived data using the same UK dataset. The same clustering algorithms and validity indexes are then applied to this dataset. In addition, the results are compared with the baseline obtained from the average daily load profiles in the first step. 

\subsection{Assessing the Results}
\label{validation}

To assess the consistency of clustering solutions, the different arrangements of households into clusters are compared. The consistency of the clusters obtained from the different clustering algorithms is used as a measure of the quality of the results with more consistency between the results suggesting a more useful method of identifying the clusters.

Measuring consistency across the clustering results using the different sets of data (load profiles and motifs) may be criticised as not necessarily providing a true measure of quality as clustering results may be consistent but not necessarily represent useful, ``true'' clusters within the data. 

The Rand index compares the different pairs of samples (i.e., each possible pair of households) and assesses the number in which each pair are in the same partition in the 2 different clustering solutions, the number where each member of the pair are in different partitions in both solutions, and the case where the members are in the same partition in one solution but a different partition in the other solution. The corrected Rand index \cite{hubert1985comparing} builds on the original work but adjusts the calculated value for the expected matching that would occur in a random arrangement. The corrected Rand index ranges from -1 to 1 with a higher value signifying better agreement between the partitions and hence a better solution.

\section{Empirical Evaluation}

\subsection{Data Selection}

A subset of the data is extracted for the peak period of 4pm to 8pm and for working days from Spring (March, April and May) 2011. Working days are weekdays excluding Scottish public holidays. Not all households have a full set of meter readings and those with less than 4 days of valid readings are excluded. The dataset has around 440,000 individual meter readings from 204 households. 

The activities of interest within a household are related to switching appliances on or off (e.g., the use of electrical appliances in cooking) and it is the changes in the readings, rather than the absolute readings, that are of most interest and are used as the basis for analysis when using motifs.

\subsection{Clustering Using the Load Profile Data}
\label{load_cluster}

The data for the evening peak period (4pm to 7.55pm) are averaged to create a representative load profile for each household. For example, all the readings for 4pm for the household are averaged to create a representative reading for 4pm, similarly for 4:05pm, etc. The 204 representative profiles, each with 48 attributes (one for each time point), are then normalised within the 0 to 1 range and used with a variety of clustering algorithms. 

\subsection{Non-Motif Variability Clustering}

Various different measures of variability of behaviour within the household can be defined without the use of motifs (e.g., \cite{dent2012finding}) and two methods are considered.

One approach is to consider the time at which the maximum usage occurred on each day during the period of analysis. These times are then used to calculate the standard deviation of the time around the mean for each household. 

\begin{figure}[h]
\centering
\includegraphics[width=0.95\columnwidth]{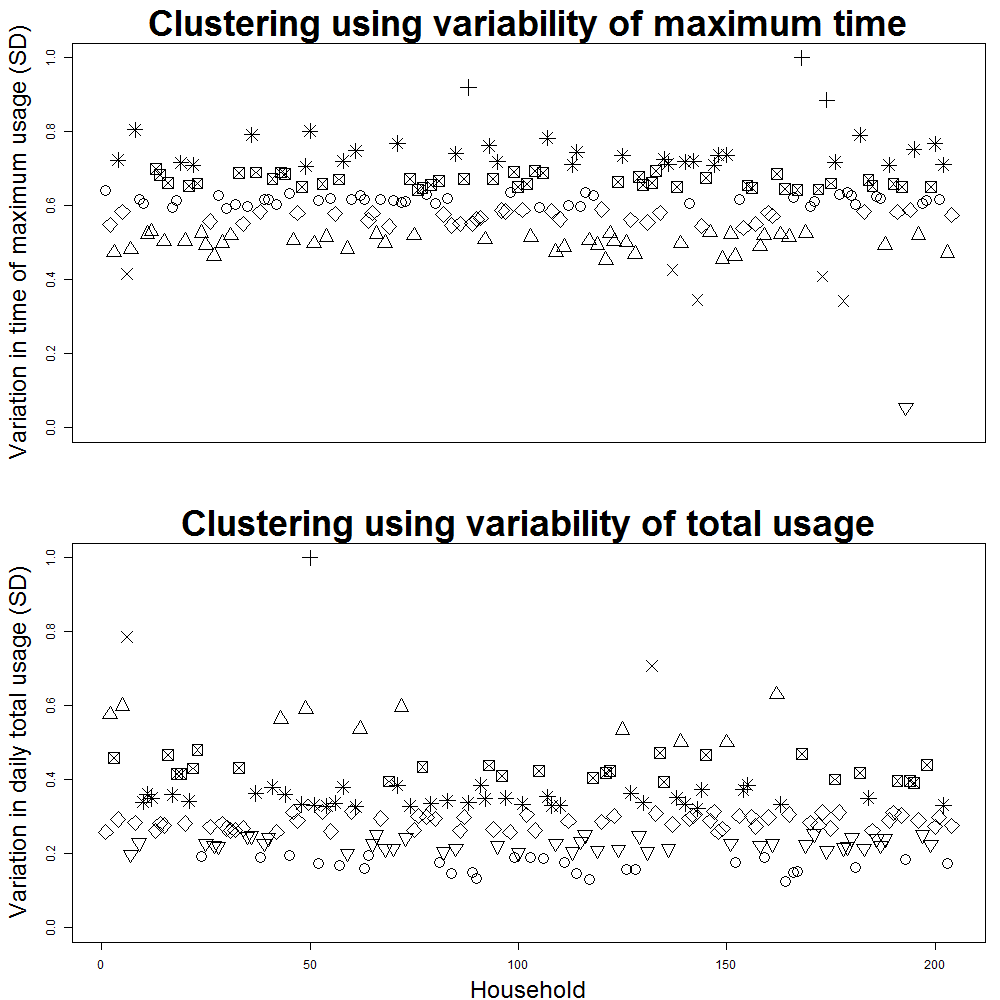}
\caption{Results for alternative variability measures}
\label{fig:other}
\end{figure}

A second approach is to consider the total usage during the peak period on each day during Spring 2011. The standard deviation of the total per day around the mean total per day also provides a measure of variability of behaviour. Each of the 2 measures are calculated and used as the basis of simple clustering using kmeans ($k=8$). The households in each of the clusters are shown in Fig. \ref{fig:other}.

There is little correspondence between the cluster assignments for the 2 methods. The corrected Rand index of 0.01 shows no correspondence beyond that expected by chance. Furthermore, there is little correspondence with the clusters obtained from the motif variability approach (detailed below). A Spearman's rho value of 0.23 shows there is little correlation between the two different variability measures. 

It is therefore concluded that neither of the non-motif measures give a useful, consistent measure of the variability of each household.

\subsection{Clustering Using the Motif Data}

This paper finds the motifs in the stream of meter data and then examines how the times of these repeating patterns vary from day to day within a household. Furthermore, the number of times a pattern repeats within a household is also used as an indication of the variability of behaviour of that household.

The motifs in the data are discovered and the attributes detailed in Section \ref{motifs} are generated. The same clustering algorithms as used for the load profile clustering are then applied to produce 8 archetypal clusters.

\subsection{Results}

Various measures that represent the variability of behaviour can be constructed and this paper considers the variability in time of maximum usage and the variability in total usage. However, as each measure is intended to represent the same thing (i.e., the variability of behaviour), the fact that there is little correlation between the measures, or the membership of the clusters generated using the measures, means that they provide a poor representation of the characteristic.

Comparing the load profile results with the motif variability results, Table \ref{tab:cluster} shows, for each of the clustering algorithms used and for each set of data, the sizes of the partitions in the solution and the values for the MIA and CDI cluster validity indexes (lower is better). 

\begin{table} [!ht]
\setlength{\tabcolsep}{4pt}
\caption{Clustering Results and Validity indexes}
\label{tab:cluster}  
\begin{center}    
\begin{tabular}{llll|lll}
\hline\noalign{\smallskip}
& \textbf{Load Profiles} &&& \textbf{Motifs} \\
\noalign{\smallskip}\hline\noalign{\smallskip}
& Cluster sizes & MIA & CDI  & Cluster sizes & MIA & CDI \\
\noalign{\smallskip}\hline\noalign{\smallskip}
  Kmeans & 10,16,19,20,20,27,44,48 & 0.593 & 1.34 & 2,5,7,26,29,37,41,57 & 0.445 & 0.641\\ 
  Fuzzy & 14,17,20,23,23,23,35,49 & 0.679 & 2.14 & 12,15,19,26,28,30,34,40 & 0.551 & 2.084\\ 
  SOM & 13,15,16,20,28,31,36,45 & 0.595 & 1.337 & 2,5,24,25,28,29,40,51 & 0.451 & 0.733\\ 
  Hier & 9,10,13,20,22,37,43,50 & 0.61 & 1.386 & 2,3,5,26,31,34,40,63 & 0.46 & 0.64 \\ 
  RF & 14,18,19,28,29,29,30,37 & 0.794 & 1.131 & 18,18,19,21,25,32,35,36 & 0.628 & 1.34\\  
\noalign{\smallskip}\hline
\end{tabular}
\end{center}
\end{table}

The MIA and CDI values show that the kmeans and SOM techniques produce similar quality solutions using the load profiles. The hierarchical algorithm is less good with the Fuzzy Cmeans algorithm being significantly poorer. The random forest and pam combination provides a good result for the CDI measure but scores poorly on the compactness of the clusters (as measured by MIA).

When using the motif variability data, the kmeans, SOM and hierarchical algorithms produce similar quality results with the Fuzzy Cmeans algorithm again producing poorer results. The random forest and pam combination provides middling results.

The MIA and CDI validity index calculations are not comparable between datasets due to the different number of attributes used. 

\begin{table}
\small
\caption{Modified Rand index of clusters using different clustering algorithms}
\label{tab:rand}       
\begin{center}
\begin{tabular}{l | lllll | lllll}
\hline\noalign{\smallskip}
& \textbf{Profiles} &&&&& \textbf{Motifs}\\
\noalign{\smallskip}\hline\noalign{\smallskip}  
& Kmeans & Fuzzy & SOM & Hier & RF & Kmeans & Fuzzy & SOM & Hier & RF \\
\noalign{\smallskip}\hline\noalign{\smallskip}
  Kmeans & 1 & 0.544 & 0.629 & 0.668 & 0.251 & 1 & 0.592 & 0.794 & 0.622 & 0.358 \\ 
  Fuzzy & 0.544 & 1 & 0.562 & 0.491 & 0.355 & 0.592 & 1 & 0.626 & 0.511 & 0.447 \\ 
  SOM & 0.629 & 0.562 & 1 & 0.49 & 0.287 & 0.794 & 0.626 & 1 & 0.591 & 0.33 \\ 
  Hier & 0.668 & 0.491 & 0.49 & 1 & 0.272 & 0.622 & 0.511 & 0.591 & 1 & 0.312 \\ 
  RF & 0.251 & 0.355 & 0.287 & 0.272 & 1  & 0.358 & 0.447 & 0.33 & 0.312 & 1\\ 
\noalign{\smallskip}\hline
\end{tabular}
\end{center}
\end{table}

Table \ref{tab:rand} gives information on the consistency of the cluster partitions as the clustering algorithm changes. The results for the Rand index show that the values are consistently closer to 1 in the case of the clusters built using motif variation. The mean values for the Rand index (after omission of the values on the diagonal) are 0.4549 for the load profiles and 0.5183 for the motif variability approach. This shows a more consistent set of partitions are created when using the motif variability than the partitions created using the load profile information. 

\begin{figure}
\centering
\includegraphics[width=0.95\columnwidth]{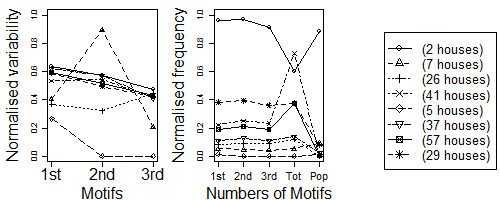}
\caption{kmeans clusters using motif variability}
\label{fig:motif_results}
\end{figure}

The results from the kmeans algorithm using the motif variability data can be seen at Fig. \ref{fig:motif_results}. The cluster with 26 houses shows very little variability in the timing of their regular activities and can be assumed to be ``creature of habit'' households who may not respond well to an incentive to change behaviour. The 2 house and the 29 house clusters show lots of repeating activities and may be best to target for interventions as there are likely to be many activities that often repeat and that may be modifiable.

\begin{figure}
\centering
\parbox{0.32\columnwidth}{
\includegraphics[height=4cm]{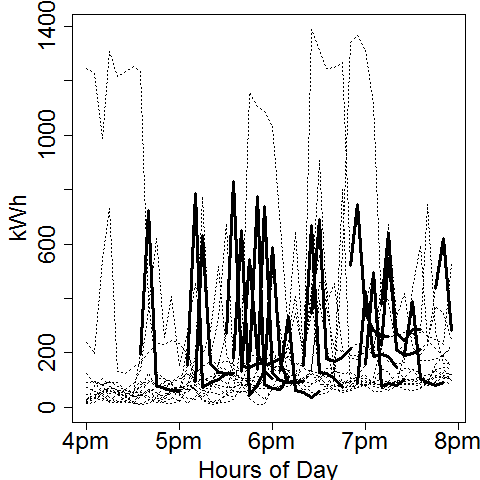}
\caption{Example house (high variability)}
\label{fig:variable_motifs}
} 
\parbox{0.32\columnwidth}{
\centering
\includegraphics[height=4cm]{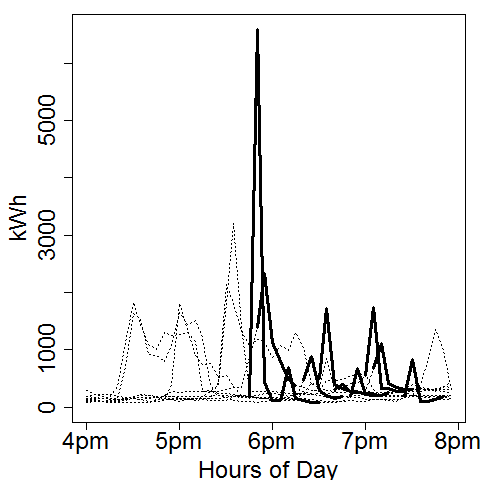}
\caption{Example house (low variability)}
\label{fig:fixed_motifs}
 }
\parbox{0.32\columnwidth}{
\centering
\includegraphics[height=4cm]{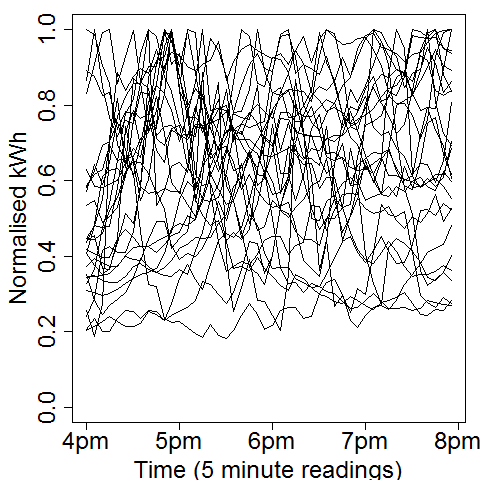}
\caption{Load profiles for high variability cluster}
\label{fig:load_profiles}
}
\end{figure}

Examining the 29 house cluster in more detail, Fig. \ref{fig:variable_motifs} show the motifs found for one of the houses and how the time of occurrence of the motif varies across the 4pm to 8pm period. In contrast, the motifs for one of the houses in the 26 house cluster are shown in Fig. \ref{fig:fixed_motifs} and the timings can be seen to be less variable.

As a comparison, the average load profiles for each of the households in the 29 house cluster are shown at Fig. \ref{fig:load_profiles}. There is little similarity between the households and hence, using the load profile shapes as the basis, little likelihood of the households being clustered together. However, the variability in timing of the motifs can be used as a method for selecting appropriate households to target and allows groupings to be designated as high or low variability.

\section{Significance and Impact}

The ability to cost effectively partition domestic households into a few meaningful archetypes based on the household electricity usage is an important problem for the electricity industry. Identifying a few archetypal representations of households is essential for cost effective implementation of DSM techniques which itself is necessary to allow the electricity industry to meet the upcoming challenges. Producing more consistent and more descriptive archetypes than currently possible will allow the deployment of effective behaviour modification interventions.

Previous work does not incorporate any measure of the variability of regular behaviour when clustering households. The variability is an important characteristic as one of the major uses of the results is to target incentives for households to vary their behaviour to provide benefit to the electricity network.

The results presented show that the ``variability in timing of motifs'' approach produces more consistent clusters across different clustering algorithms compared to the consistency of clustering using just the daily load profiles. 

The symbolisation technique is effective in detecting repeating patterns (motifs) that are approximately the same shape. Depending on the type of intervention planned for a subset of the households (for example, incentives to change overall electricity usage from day to night, or to influence short periods of usage during the peak period), different sizes of motifs may be used.

This work shows a novel approach to using electricity meter data to cluster households that enhances and complements the existing techniques based on the daily load profiles. 

\section*{Acknowledgements}
This work was possible thanks to RCUK Energy Programme and EPSRC grant references EP/I000496/1 and EP/G065802/1 and forms part of the Desimax project \cite{Kiprakis2011}.

Thanks are due to Pavel Senin for providing R code implementing the SAX method.

\bibliographystyle{splncs}

\bibliography{/Users/ird/Documents/Jabref_database/References}

\end{document}